\def\year{2022}\relax
\documentclass[letterpaper]{article} 
\usepackage{aaai22}  
\usepackage{times}  
\usepackage{helvet}  
\usepackage{courier}  
\usepackage[hyphens]{url}  
\usepackage{graphicx} 
\usepackage{natbib}  
\usepackage{caption} 
\DeclareCaptionStyle{ruled}{labelfont=normalfont,labelsep=colon,strut=off} 
\usepackage{array}
\usepackage{booktabs}
\usepackage{amsmath,amsfonts,amssymb,amsthm}
\newtheorem{theorem}{Theorem}
\newtheorem{lemma}{Lemma}

\usepackage{xcolor}
\usepackage{multirow}

\DeclareMathOperator*{\arginf}{arginf}
\newcommand{\system}{mixer}
\newcommand{\sys}{mix}
\newcommand{\MGM}{MGM}

\usepackage[ruled,vlined,linesnumbered]{algorithm2e}
\usepackage{minitoc}
\usepackage[toc,page,header]{appendix}

\usepackage{bibunits}
\defaultbibliography{references}

\title{Multiscale Generative Models: Improving Performance of a Generative Model Using Feedback from Other Dependent Generative Models}
\author{
Changyu Chen,\textsuperscript{\rm 1}
Avinandan Bose,\textsuperscript{\rm 2}
Shih-Fen Cheng,\textsuperscript{\rm 1}
Arunesh Sinha\textsuperscript{\rm 1}
}
\affiliations{
    \textsuperscript{\rm 1} Singapore Management University\\
    \textsuperscript{\rm 2} Indian Institute of Technology Kanpur\\
    cychen.2020@phdcs.smu.edu.sg,
    avibose@iitk.ac.in,
    sfcheng@smu.edu.sg,
    aruneshs@smu.edu.sg
}

\begin{document}

\maketitle

\begin{abstract}
Realistic fine-grained multi-agent simulation of real-world complex systems is crucial for many downstream tasks such as reinforcement learning. Recent work has used generative models (GANs in particular) for providing high-fidelity simulation of real-world systems. However, such generative models are often monolithic and miss out on modeling the interaction in multi-agent systems. In this work, we take a first step towards building multiple interacting generative models (GANs) that reflects the interaction in real world. We build and analyze a hierarchical set-up where a higher-level GAN is conditioned on the output of multiple lower-level GANs. We present a technique of using feedback from the higher-level GAN to improve performance of lower-level GANs. We mathematically characterize the conditions under which our technique is impactful, including understanding the transfer learning nature of our set-up. We present three distinct experiments on synthetic data, time series data, and image domain, revealing the wide applicability of our technique. 
\end{abstract}

\section{Introduction}
Realistic simulations are an important component of training a reinforcement learning based decision-making system and transferring the policy to the real world. While typical video game-based simulations are, by design, realistic, the same is very difficult to claim when the problem domain is a real-world multi-agent system such as stock markets and transportation. Indeed, recent works have used generative models to design high-fidelity simulators using real world data~\cite{li2020generating,sun2020decision,shi2019virtual,chen2019generative}, supposedly providing a higher degree of fidelity than traditional agent-based simulators. However, such generative models do not provide a fine-grained simulation of each agent in a multi-agent system, instead modeling the system as a single generative model. A fine-grained simulation can be informative in the design of interventions to optimize different aspects of the problem, e.g., how traders react to a new stock market policy. In this work, we take a first step towards building multiple interacting generative models (specifically, GANs), one for each type of agent, providing basic building blocks for a fine-grained high-fidelity simulator for complex real-world systems. 

To the best of our knowledge, no prior work has built multi-agent simulations with multiple interacting generative models. 
Our work focuses on utilizing the interaction in a multi-agent system for better (generative) modeling of an individual agent for whom the data is limited or biased. 
Reflecting the multi-agent interaction in the real world, we set up a hierarchy of generative model instances, called Multiscale Generative Models (\MGM{}), where the higher level is a conditional model that generates data conditioned on data generated by multiple distinct generative model instances at the lower level. We particularly note that our hierarchy is among different GAN instances (with distinct generative tasks), unlike hierarchy among multiple generators or discriminators for a single generative task that has been explored in the literature~\cite{hoang2018mgan}.

Our \emph{first contribution} is an architecture of how to utilize the feedback from a higher-level conditional GAN to improve the performance of a lower-level GAN. The higher-level GAN takes as input the output of (possibly multiple) lower-level GANs, mirroring agent interaction in the real world, e.g., prices offered in multiple regions (apart from other random factors) determining the total demand of electricity. We repurpose the discriminator (or critic in WGANs) of the higher-level model to provide an additional feedback (as an additional loss term) to the lower-level generator. 

As our \emph{second contribution}, we provide a mathematical interpretation of the interacting GAN set-up, thereby characterizing that the coupling between the higher-level and the lower-level models determines whether the additional feedback improves the performance of the lower-level GANs or not. We identify two scenarios where our set-up can be especially helpful. One is where there is a small amount of data available per agent at the lower-level and another where data is missing in some biased manner for the lower-level agents. Both these scenarios are often encountered in the real-world data, for example, in the use case of modeling electricity market, the amount of demand data for a new consumer might be small and the feedback from a higher-level model about such consumer's expected behaviors helps in building a more realistic behavioral model. Moreover, the case of small amount of data can also be viewed as a few-shot generation problem.

Finally, we provide three distinct sets of experiments revealing the fundamental and broadly applicable nature of our technique. The first experiment with synthetic data explores the various conditions under which our scheme provides benefits. Our second experiment with electric market time series data from Europe explores a set-up where electricity prices from three countries (plus random factors) determine demands, where we show improved generated price time series from one of the lower-level data generator using our approach. In our third experiment, we demonstrate the applicability of our work for a few-shot learning problem where the higher-level CycleGAN is a horse to zebra convertor trained using ImageNet, and a single lower-level GAN generates miniature horse images using feedback from the CycleGAN and small amount of data. Overall, we believe that this work provides a basic ingredient of and a pathway for building fine-grained high-fidelity multi-agent simulations. All missing proofs are in appendix.

\begin{figure*}
    \centering
    \includegraphics[width=15cm,keepaspectratio]{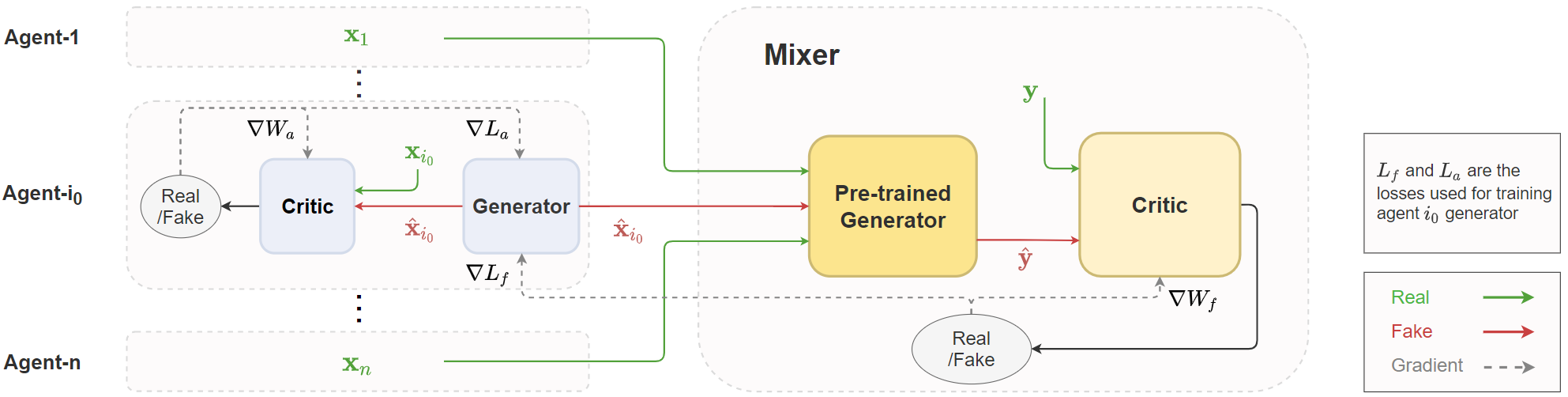}
    \caption{Architecture of the proposed MGM model.}
    \label{fig: arch}
\end{figure*}

\section{Related Work}
\textbf{Generative Models}: There is a huge body of work on generative modeling, with prominent ones such as VAEs~\cite{kingma2013auto} and GANs~\cite{goodfellow2014generative}. The closest work to ours is a set-up in federated learning where multiple parties own data from a small part of the full space and the aim is to learn one generator over the full space representing the overall distribution. Different approaches have been proposed to tackle this, where multiple discriminators at multiple parties train a single generator~\cite{hardy2019md} or multiple generators at different parties share generated data~\cite{ferdowsi2020brainstorming} or network parameters~\cite{rasouli2020fedgan} to learn the full distribution. 
Our motivation, set-up, and approach is very different. Our motivation is fine-grained multi-agent simulation where we exploit the learned knowledge of the real world interaction (embedded in the higher level conditional GAN) to improve a lower level agent's generative model. Dependency among agents is crucial for our work, whereas these other approaches have independent parties.

Other non-federated learning works also employ multiple generators or discriminators to improve learning stability~\cite{hoang2018mgan,durugkar2016generative} or prevent mode collapse~\cite{ghosh2018multi}. \citet{pei2018multi} proposed multi-class discriminators structure, which is used to improve the classification performance cross multiple domains. 
\citet{liu2016coupled} proposed to learn the joint distribution of multi-domain images by using multiple pairs of GANs. The network of GANs share a subset of parameters, in order to learn the joint distribution without paired data. Again, our main insight is to exploit dependencies among agents which is not considered in these work. 

\textbf{Few-shot Image Generation}: Few-shot image generation aims to generate  new and diverse examples whilst preventing overfitting to the few training data. Since GANs frequently suffer from memorization or instability in the low-data regime (e.g., less than 1,000 examples), most prior work follows a few-shot adaptation technique, which starts with a source model, pretrained on a huge dataset, and adapts it to a target domain. 
They either choose to update the parameters of the source model directly with different forms of regularization~\cite{wang2018transferring,mo2020freeze,ojha2021few,li2020few}, or embed a small set of additional parameters into the source model~\cite{noguchi2019image,wang2020minegan,robb2020few}.
Besides these, some works also exploit data augmentation to alleviate overfitting~\cite{NEURIPS2020karras,zhao2020differentiable} but this approach struggles to perform well under extreme low-shot setting (e.g., 10 images)~\cite{ojha2021few}. 
Different from these work, our method exploits abundant data that is dependent on the target domain data and provides effective feedback to the GAN that performs adaptation (lower level GAN), leading to higher quality generation results. Note that our \MGM{} does not have any specific requirement for the lower level GAN so it can benefit a bunch of existing work. 

\section{Preliminaries and Problem Formulation} \label{sec:prelim}
Although our approach is independent of the type of generative model used, to explain our main solution approach concretely, we adopt a well-known generative model called WGAN, which is briefly introduced below. 
Every WGAN has a generator network $G_\theta$ and a critic network $C_w$, where $\theta, w$ are weights of the networks. The generator generates samples $y = G_\theta(z, c) \in Y$ from a learned conditional distribution $\mathbb{P}^g_{y\vert c}$, where $z$ is some standard noise (e.g., Gaussian) and $c \in C$ is a condition. The goal is to learn the true distribution $\mathbb{P}^r_{y\vert c}$.  In an unconditional WGAN, the critic simply outputs an estimate of the Wasserstein-1 distance, $W(\mathbb{P}^r, \mathbb{P}^g)$, between the real and generated distributions, given samples from each. For the conditional case, the condition $c$ forms part of the input $(y,c)$ for the critic. Let $\mathbb{P}_{joint}^r$ denote the distribution over $Y \times C$ with $c \sim \mathbb{P}^r_c$ and $y \sim \mathbb{P}^r_{y\vert c}$, and similarly $\mathbb{P}_{joint}^g$ denotes the distribution with $c \sim \mathbb{P}^r_c$ and $y \sim \mathbb{P}^g_{y\vert c}$. With the condition as input, the critic estimates the distance between joint distributions, $W(\mathbb{P}_{joint}^r, \mathbb{P}_{joint}^g)$. The expected loss for the generator is $- E_{z, c \sim \mathbb{P}^r_c}[C_{w}(G_\theta(z,c)]$ and for the critic is\footnote{As a heuristic, generated sample is mixed with real sample in WGAN-GP; we ignore this in writing for ease of presentation.}
\begin{align*}
  E_{(y,c) \sim \mathbb{P}^g_{joint}}[C_{w}(y,c)] - & E_{(y,c) \sim \mathbb{P}^r_{joint}}[C_{w}(y,c)] + \\
  & \qquad 
  \lambda \mathcal{J}_{C_w,\mathbb{P}^g_{joint}}  
\end{align*}
where $\mathcal{J}$ is the gradient penalty regularizer in WGAN-GP~\cite{gulrajani2017improved} with weight $\lambda$. The actual loss optimized is an empirical version of the above expectations, estimated from data samples. A subtle implementation issue (unaddressed in prior work) arises with the $\mathcal{J}$ term when the condition is continuous; we discuss how we address this in the appendix.



\textbf{Problem Formulation}: Let there be $n$ agents, where the agent in $i^{th}$ position produces data distributed in the space $X_i \subset \mathbb{R}^{d_i}$ for some positive integer dimension $d_i$. Corresponding to position $i$, let $\mathbb{P}^r_i$ be a probability distribution and the agent generates $x_i \in X_i$ distributed as $x_i \sim \mathbb{P}^r_i$. Denote the data generated by all the agents as a single vector $x = \langle x_1, \ldots, x_n \rangle$. These lower-level agents generate data that feeds into a higher-level agent, which we call as \emph{\system{}} to distinguish it from the lower-level agents.
The \system{} generates data in space $Y \subset \mathbb{R}^{d}$ for some positive integer dimension $d$ with a conditional distribution $\mathbb{P}^{r}_{y \vert x}$. A pictorial representation of the set-up is shown in Figure~\ref{fig: arch}. 

Our goal is to obtain a fine-grained model of the agents as well as the \system{}. With sufficient data, building multiple GAN models for the \system{} and every agent is not very hard. However, the data for some lower-level agents are often limited or even corrupted; e.g., when an agent has \emph{newly arrived} into the system. For example, this could happen with a new electricity supplier in the electricity market, or a new trader in a stock market, or even a new image generation task with low or biased amount of data. 
Thus, we assume one of the $n$ agents, indexed by $i_0$ has newly arrived and replaces the prior agent at index $i_0$. 
This leads to the technical problem that we solve in this paper.

\textbf{Problem Statement}: Our goal is to utilize the higher-level \system{} GAN to improve upon the generative model of a lower-level agent GAN, namely one that is newly arrived. 

We believe that the above is tractable since the \system{} model is trained on and relies on the output distribution of every kind of agent, and hence encodes information about the agent-level model within its own model. This information can be used to guide the training of an agent-level GAN.

\textcolor{black}{\textbf{Examples}: We provide two possible real world practical applications of a solver of the aforementioned problem. This is apart from the practical few-shot learning experiment we present later. First, a biased data scenario. Consider three transportation network companies (TNCs), such as Uber, providing online ride-hailing services in a city with multiple districts. They would like to model the average behavior of drivers in different districts to help with optimized vehicle dispatching. This can be done by training a generative model for every district (each district as an agent) if a TNC has all the data. However, a TNC only has part of the drivers’ data, that is, a biased dataset. E.g., 30\% drivers work for TNC-1, while 40\% and 30\% work for TNC-2 and TNC-3 respectively. Due to the competitive relationship, TNCs do not share data with each other. They agree on providing their data to a third party, where all the data is used to pre-train a mixer that models a city-level output (e.g., total supply). When a TNC trains an agent, the agent can obtain extra feedback by feeding its own generator’s output to the third part (mixer) and receiving the gradient feedback (as explained in the approach section). During the process, the third party does not share any data with a TNC.}

\textcolor{black}{Next, consider a low data scenario. Continuing from the above example, consider a new TNC (a fourth TNC) that starts to provide service in the city. As it is a new TNC, it has very limited data. In this case, the fourth TNC can also improve its own model by the feedback from the mixer.}

\section{Approach}
Our approach, which we call \MGM{}, to the problem stated above relies upon the architecture in Fig.~\ref{fig: arch}. We choose GANs as our generative model (different types of GANs for different problems). As shown in the architecture, the \system{} GAN is a conditional GAN in which the conditions are the output of the agent GANs. We use the notation: $x_{-i_0}$ as a vector in $\times_{j \neq i_0} X_j$ and $\langle x_{-i_0}, x_{i_0} \rangle$ denotes the full vector with the $i_0$ component. 

The $i^{th}$ agent generator takes as input noise $z_i$ to produce a sample $x_i = G_{\theta_i}(z_i)$ from its learned distribution. The \system{} generator takes as input (1) noise $z_{\sys{}}$ and (2) the output $x = \langle x_1, \ldots, x_n \rangle$ of the agent generators as a  condition to produce a sample $y = G_{\theta_{\sys{}}}(z_{\sys{}}, x)$ from its learned distribution.
The conditional \system{} GAN is trained in the traditional manner using available data for all the agents. 
We assume that there is plenty of data to train the \system{} GAN effectively. That is, the learned \system{} generator $G_{\sys{}}$ models the distribution $\mathbb{P}^{r}_{y \vert x}$ accurately. 

We focus on the training of one \emph{newly arrived} agent $i_0$, which replaces the prior agent $i_0$. The position $i_0$ has a true average distribution $\mathbb{P}^r_{i_0}$. This new arrival happens after the \system{} GAN is trained. $D^r_y \subset Y$ is the real output data corresponding to the mixer. We also have datasets for every agent $j \neq i_0$, given by $D^r_j \subset X_j$ and $D^r_{i_0} \subset X_{i_0}$. In particular, $D^r_{i_0}$ has low amount of or biased data that prevents learning the true distribution $\mathbb{P}^r_{i_0}$ just using $D^r_{i_0}$.

Our technical innovation is in training generator of agent $i_0$. The framework for this is shown in Algorithm~\ref{alg:framework}. The agent trainer has access to the datasets stated above, $D^r_y, D^r_j$ for all agents $j$, as well as the pre-trained $G_{\sys{}}$ of the \system{}. The agent $i_0$ generator is trained using an empirical version $L_a$ (i.e., sample average estimate) of its own expected loss $\mathcal{L}_a$ and also using an empirical version $L_f $ of the feedback $\mathcal{L}_{f}$ from the \system{} critic (line 16-17). There are two ways to combine these loss terms: (1) form one combined loss as $\alpha L_a + (1-\alpha) L_f$ where $\alpha \in (0,1)$ is a tunable hyperparameter and (2) train with $L_a$ for some iterations and then $L_f$ for some iterations (alternate updating). In experiments, we use the former for synthetic and time series data, and the latter for image domain.

The empirical estimates for $\mathcal{L}_a, \mathcal{L}_f$ are obtained using samples in line 14-15. Additionally, $\mathcal{L}_a, \mathcal{L}_f$ depend on the agent level critic $C_{w_{i_0}}$ and \system{} level critic $C_{w_{\sys{}}}$ respectively (see example below). $C_{w_{i_0}}$ and $C_{w_{\sys{}}}$ are trained with the empirical form $W_a$ and $W_f$ of the expected loss $\mathcal{W}_a $ and $\mathcal{W}_f $, respectively (line 10 and 12). These empirical losses $W_a$ and $W_f$ are built from the samples in line 5-9.

\textbf{Concrete Example}: We use the WGAN-GP loss terms here. With WGAN-GP, the loss $\mathcal{L}_a$ and $\mathcal{L}_{f}$ would be $- E_{x_{i_0} \sim \mathbb{P}^g_{x_{i_0}}}[C_{w_{i_0}}(x_{i_0})]$ and $- E_{y \sim \mathbb{P}^g_y}[C_{w_{\sys{}}}(y)]$, respectively. $\mathcal{W}_a $ is the standard WGAN-GP loss
\begin{align*}
E_{x_{i_0} \sim \mathbb{P}^g_{x_{i_0}}}[C_{w_{i_0}}(x_{i_0})] - E_{x_{i_0} \sim \mathbb{P}^r_{x_{i_0}}}[C_{w_{ i_0}}(x_{i_0})] + \\
\qquad \lambda \mathcal{J}_{C_{w_{i_0}},\mathbb{P}^g_{x_{i_0}}}
\end{align*}
For $\mathcal{W}_f$, we first introduce some notation. Let $\mathbb{P}^r_y$ be the real marginal distribution over $Y$, marginalized over all agents distributions. Let $\mathbb{P}^g_y$ be the marginal generated distribution over $Y$ formed by the following process: $x_j \sim \mathbb{P}^r_j$ for $j \neq i_0$, $x_{i_0} \sim \mathbb{P}^g_{i_0}$, $y \sim \mathbb{P}^r_{y \vert x}$. Then, $\mathcal{W}_f$ is
$$
E_{y \sim \mathbb{P}^g_y}[C_{w_{\sys{}}}(y)] - E_{y \sim \mathbb{P}^r_y}[C_{w_{\sys{}}}(y)]  + \lambda_{\sys{}} \mathcal{J}_{C_{w_{\sys{}}},\mathbb{P}^g_{y}}
$$
In the above expressions, all expectations are replaced by averages over samples to form the counterpart empirical losses. These samples are obtained as follows: $D^r_y$ are samples from $\mathbb{P}^r_y$ (line 9) and samples from $\mathbb{P}^g_y$ are generated by (1) forming $x$, by obtaining $x_j$ from $ D^r_j$ for $j \neq i_0$ (line 6), obtaining $x_{i_0}$ by running generator for $i_0$ (line 7), and (2) finally obtaining $y$ by running the pre-trained \system{} generator (which is assumed to be accurate) conditioned on $x$ (line 8).

We reiterate that Algorithm~\ref{alg:framework} is a framework, and in the above paragraph we provide a possible instantiation with WGAN loss. The framework can be instantiated with CycleGAN and StyleGAN as well as a time series generator COT-GAN. These are used in the experiments, and we provide the detailed losses in the appendix due to space constraint.
\begin{algorithm}[t]
\DontPrintSemicolon
\caption{\textit{\MGM{} Template}}  \label{alg:framework}
\textbf{Inputs}: Pre-trained $G_{\sys{}}$, datasets $D^r_y$, $D^r_j \;  \forall j$,\; 
$N(0,I)$ is standard normal dist., $U(\cdot)$ is uniform dist.\;
\While{$\theta_{i_0}$ has not converged}{
\For{$m$ times}{
 Sample $\{z_k\}_{k=1}^K, \{z'_k\}_{k=1}^K \sim N(0,I)$  \;
 Sample $\{x^r_{j, k}\}_{k=1}^K \sim U(D^r_{j})$\;
 $x^g_{i_0, k} = G_{\theta_{i_0}}(z_k)$ , $x^g_{k} = \langle x^r_{-i_o, k}, x^g_{i_0, k} \rangle $\; 
 $y^g_k = G_{\sys{}}(z'_k, x^g_{k} )$ \;
 Sample  $\{y^r_k\}_{k=1}^K \sim U(D^r_{y})$\;
 Form loss $W_a$ using $\{x^g_k, x^r_k\}_{k=1}^K$\;
 Update $w_{i_0}$ using $\nabla_{w_{i_0}} {W}_a  $\;
 Form loss $W_f$ using $\{y^g_k, y^r_k\}_{k=1}^K$\;
 Update $w_{\sys{}}$ using $\nabla_{w_{\sys{}}} {W}_f  $\;
 }
 Sample $\{z_k\}_{k=1}^K, \{z'_k\}_{k=1}^K \sim N(0,I)$\;
 Sample $\{x^r_{j, k}\}_{k=1}^K \sim U(D^r_{j})$ \;
 Form loss $L_a$ using $G_{\theta_{i_0}}(z_k), C_{w_{i_0}}$\;
  Form loss $L_f $ using $G_{\sys{}} \big (z_k', \langle x^r_{-i_o, k}, G_{\theta_{i_0}}(z_k) \rangle ), C_{w_{\sys{}}}$\;
 Update $\theta_{i_0}$ using $ \nabla_{\theta_{i_0}} {L}_a $ and $  \nabla_{\theta_{i_0}}{L}_f  $\;
 }
\end{algorithm}
\subsection{Mathematical Characterization}
We derive mathematical result for WGAN. As the high level idea of all GANs are similar, we believe that the high level insight obtained at the end of sub-section hold for any type of GAN. From the theory of WGAN~\cite{arjovsky2017wasserstein}, the loss term $\mathcal{L}_f$ estimates the Wasserstein-1 distance between $\mathbb{P}^r_y$ and $\mathbb{P}^g_y$. As these two distributions differ only in the way $x_{i_0}$ is generated, it is intuitive that $\mathcal{L}_f$ provides meaningful feedback. However, there are many interesting conclusions and special cases that our mathematical analysis reveals, which we show next.\footnote{We work in Euclidean spaces and standard Lebesgue measure. Probability distribution refers to the probability measure function.}

\begin{figure*}
    \centering
    \includegraphics[width=14cm,keepaspectratio]{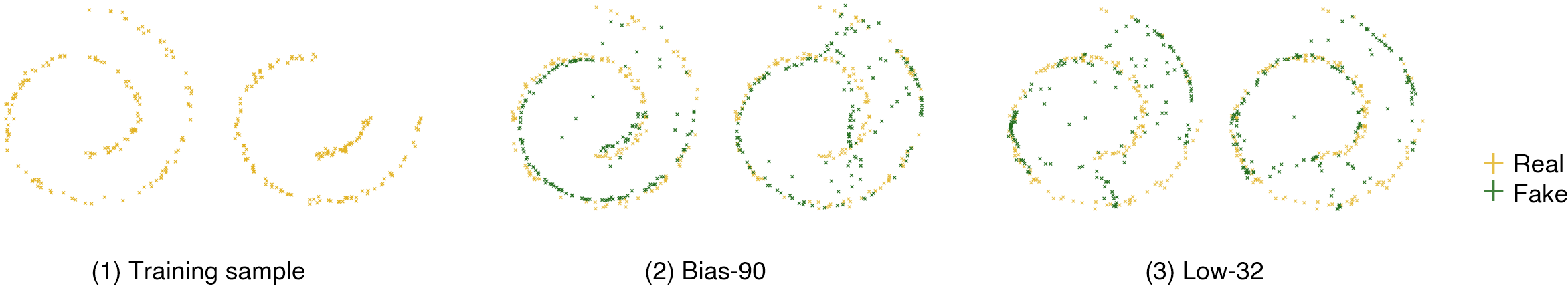}
    \caption{(1) Training sample: left fig. is full training data, right fig. shows missing data in the top right area (biased data); (2) given biased data, left fig. shows the baseline, right fig. shows WGAN-GP-C; (3) given low data (32 samples), left fig. shows the baseline, right fig. shows WGAN-GP-C. Note that all results shown here use $\beta=0.7$.}
    \label{fig: exp1-summary}
\end{figure*}

\textbf{Background on Wasserstein-1 metric}: Let $\Pi_{\mathbb{P}_i,\mathbb{P}_j}$ be the set of distributions (also called couplings) whose marginal distributions are $\mathbb{P}_i$ and $\mathbb{P}_j$, i.e., $\int_{X_i \times A} d\pi(x_i,x_j)  = \mathbb{P}_j(A)$ and $\int_{A \times X_j } d\pi(x_i,x_j)  = \mathbb{P}_i(A)$ for any event $A$ in the appropriate $\sigma$-algebra, for any $\pi \in \Pi_{\mathbb{P}_i,\mathbb{P}_j}$. The Wasserstein-1 metric is:
\begin{align}
W(\mathbb{P}_i,\mathbb{P}_j) = \inf_{\pi \in \Pi_{P_i,P_j}} \int^{}_{X_i \times X_j} d(x_i,x_j) \; d\pi(x_i,x_j)  \label{eq:primal}
\end{align}
It is known that the Wasserstein-1 metric is also equal to
$$
\sup_{\substack{f(x_i) + g(x_j) \leq d(x_i,x_j) \\
f\in \mathcal{C}_b(X_i), g \in \mathcal{C}_b(X_j)}} \int^{}_{X_i } f(x_i) \; d \mathbb{P}_i(x_i)  + \int^{}_{X_j } g(x_j) \; d \mathbb{P}_j(x_j) 
$$
where $\mathcal{C}_b(X)$ is the space of bounded and continuous real valued functions on $X$. 
The functions $f$ and $g$ are also the dual functions~\cite{villani2009optimal} as they correspond to dual variables of the infinite linear program that Eq.~\ref{eq:primal} represents. It is known that the minimizer of Eq.~\ref{eq:primal} and maximizer functions of the dual form exists~\cite{villani2009optimal}. 

\textbf{Analysis of $\mathcal{L}_f$}: We are given the system generator that represents the distribution $\mathbb{P}^{r}_{y|x}$, with probability density function given by $P^{r}_{y|x}(\cdot)$. We also have true distributions $\mathbb{P}^r_j$ for all $j$.  The data that we are interested in generating is from $\mathbb{P}^r_{i_0}$, 
and the corresponding agent level generator generates data from $\mathbb{P}^g_{i_0}$. We aim to have $\mathbb{P}^g_{i_0}$ close to $\mathbb{P}^r_{i_0}$. Towards this end, we repurpose the critic of the \system{} GAN. In training, the critic is fed real data that is distributed according to the density $P^r_{y}(y) = \int_{X} P^r_{joint}(y,x) dx$ where 
$ P^r_{joint}(y,x) =  P^{r}_{y|x}(y|x) \times \prod_j P^r_j(x_j)$
. The \system{} critic is additionally fed generated data that is distributed according to the density 
$P^g_{y}(y) = \int_{X} P^g_{joint}(y,x) dx$ where
$P^g_{joint}(y,x) = P^{r}_{y|x}(y|x) \times P^g_{i_0}(x_i) \times \prod_{j \neq i_0} P^r_j(x_j)$. 
Thus, the repurposed \system{} critic estimates the Wasserstein-1 distance between $\mathbb{P}^r_{y}(\cdot)$ and $\mathbb{P}^g_{y}(\cdot)$ given as
$$
W(  \mathbb{P}^r_y, \mathbb{P}^g_y)  = \inf_{\pi \in \Pi_{\mathbb{P}^r_{y}, \mathbb{P}^g_{y}}}\int_{Y\times Y} d(y, y') \; d\pi(y, y')
$$
In order to analyze the above feedback from the \system{} to the agent generator, we define some terms. Consider
$$
W(\mathbb{P}^{r}_{y|\hat{x}}, \mathbb{P}^{r}_{y|\hat{x}'})  = \inf_{\pi \in \Pi_{\mathbb{P}^{r}_{y|\hat{x}},\mathbb{P}^{r}_{y|\hat{x}'}} }
\int^{}_{Y \times Y} d(y,y') \; d\pi (y,y') 
$$
Dual functions $f^*_{\hat{x}}$ and $g^*_{\hat{x}'}$ exist here such that $f^*_{\hat{x}}(y) + g^*_{\hat{x}'}(y') \leq d(y,y')$ for all $y,y'$ as well as for any $\hat{x},\hat{x}'$ and

\begin{align*}
W&(\mathbb{P}^{r}_{y|\hat{x}}, \mathbb{P}^{r}_{y|\hat{x}'})  = \\
&\int^{}_{Y } f^*_{\hat{x}}(y)  P^r_{}(y) \; d y  + \int^{}_{Y } g^*_{\hat{x}'}(y) P^g_{}(y) \; d y 
\end{align*}

Let $\pi^*_d(x_j, x_j')$ denote the distribution (or coupling) that achieves the minimum for the Wasserstein-1 distance below between the same distributions $\mathbb{P}^r_{i_0}$ and $\mathbb{P}^g_{i_0}$ 
$$
\pi^*_d =  \arginf\limits_{\pi \in \Pi_{\mathbb{P}^r_{i_0},\mathbb{P}^g_{i_0}}} \int^{}_{X_{i_0} \times X_{i_0}} d(x_{i_0},x'_{i_0}) \; d\pi(x_{i_0},x_{i_0}')
$$
Note that $\pi^*$ takes $d$ as a subscript, i.e., it depends on the distance metric $d$. We denote this explicitly since we later deal with different distance metrics. We will also use subscript $d$ for $W(\cdot, \cdot)$, if not obvious. $\otimes_{j \neq i_0} \mathbb{P}^r_j$ denotes the joint probability distribution (product measure) over $\times_{j \neq i_0} X_j$. The next result provides a mathematical formulation of the Wasserstein-1 distance feedback provided from the mixer critic.

\begin{theorem} \label{thm:main}
Assume that (1) all spaces $X_j$'s, $Y$ are compact, (2) all probability density functions are continuous, bounded and $> 0$ over its domain, and (3) $f^*_{x}(y), g^*_{x}(y)$ are jointly continuous in $x,y$.  Then, the following holds:
\begin{align*}
W( & \mathbb{P}^r_y, \mathbb{P}^g_y)  = \int_{X_{i_0} \times X_{i_0}} E_{x_{-i_0} \sim \otimes_{j \neq i_0} \mathbb{P}^r_j}\Big[ \\
& \qquad\quad W(\mathbb{P}^{r}_{y| \langle x_{-i_0}, x_{i_0} \rangle}, \mathbb{P}^{r}_{y|\langle x_{-i_0}, x'_{i_0} \rangle}) \Big]   \; d\pi^*_{d}(x_{i_0}, x_{i_0}'), 
\end{align*}
where $d$ is any metric on $X_{i_0}$.
\end{theorem}
We provide further explanation of the above mathematical result. The following result aids in the explanation:
\begin{lemma} \label{lemma:dist}
Define $M_{x_{-i_0}}(\cdot )$ as the function $M_{x_{-i_0}}(x_{i_0}) = \mathbb{P}^{r}_{y| \langle x_{-i_0}, x_{i_0} \rangle }$ and assume all conditions in Theorem~\ref{thm:main} hold. 
If $M_{x_{-i_0}}(\cdot )$ is an injection for all $x_{-i_0}$, then 
\begin{align*}
d_0&(x_{i_0}, x'_{i_0}) = \\
& E_{x_{-i_0} \sim \otimes_{j \neq i_0} \mathbb{P}^r_j}\big[ W(\mathbb{P}^{r}_{y| \langle x_{-i_0}, x_{i_0} \rangle}, \mathbb{P}^{r}_{y|\langle x_{-i_0}, x'_{i_0} \rangle }) \big ]
\end{align*}
is a distance metric on $X_{i_0}$. 
If in the above case, $M_{x_{-i_0}}(\cdot )$ is not an injection then $d_0$ is a pseudo-metrics.
\end{lemma}
Using the above result, if $M_{x_{-i_0}}(\cdot )$ is an injection $\forall x_{-i_0}$, then we can check that by choosing $d\! =\! d_0$ in Thm.~\ref{thm:main} we get
$$
W( \mathbb{P}^r_y, \mathbb{P}^g_y ) = W_{d_0}(  \mathbb{P}^r_{i_0}, \mathbb{P}^g_{i_0}) 
$$
In words, under the injective assumption, the feedback $\mathcal{L}_f$ from the \system{} critic is the Wasserstein-1 distance between real and fake distributions of position $i_0$ but measured in a different underlying distance metric $d_0$ on $X_{i_0}$ induced by the mapping $M_{x_{-i_0}}(\cdot )$. 
When $M_{x_{-i_0}}(\cdot )$ is not an injection, then it induces a pseudo-metric on $X_{i_0}$. A pseudo-metric satisfies all properties of a distance metric except that two distinct points may have distance $0$. This provides less fine-grained feedback with a push to match probability density only over those points with $d_0 > 0$, and in the extreme case when $M_{x_{-i_0}}(\cdot )$ is a constant function then $W(\mathbb{P}^{r}_{y| \langle x_{-i_0}, x_{i_0} \rangle }, \mathbb{P}^{r}_{y|\langle x_{-i_0}, x'_{i_0} \rangle})$ is zero for any $x_{i_0}, x'_{i_0}$ and thus the feedback $\mathcal{L}_f$ from the \system{} critic is zero. Indeed, $M_{x_{-i_0}}(\cdot )$ constant is the case when the \system{} output distribution $\mathbb{P}^r_y$ does not depend on $x_{i_0}$ and hence the \system{} model does not encode any useful information about agent $i_0$. 

At a high level, how the \system{} depends on the agent $i_0$ (captured in $M_{x_{-i_0}}(\cdot )$) determines the nature and magnitude of the feedback. This feedback combined with the agent $i_0$'s own loss still aims to bring $\mathbb{P}^g_{i_0}$ closer to $\mathbb{P}^r_{i_0}$.

\section{Experiments}
We showcase the effectiveness of \MGM{} on three different domains\footnote{\textcolor{black}{\url{https://github.com/cameron-chen/mgm} : code and data are available at this url}}: (1) synthetic data modeling relationship between a \system{} and two agents, (2) time series generation task, and (3) few-shot image generation that reveals the transfer learning nature of the \MGM{} set-up. 
We use different specialized GANs for each domain. 

For the synthetic and image data, we ran the experiments on a server (Intel(R) Xeon(R) Gold 5218R CPU, 2x Quadro RTX 6000, 128GB RAM) on the GPU. For time series data, we ran the experiments on a cloud instance (14 vCPUs, 56GB memory, 8x Tesla K80) on the GPU. For all experiments, we use Adam optimizer with a constant learning rate tuned between 1e-5 and 1e-3 by optuna~\cite{optuna_2019}.

\textbf{Evaluation}: Our evaluation methodology has the following common approach: we first train a \system{} GAN (or use pre-trained one) using full data available. We train the agent $i_0$ GAN using low or biased data only and additionally with feedback from the \system{}. We then compare the performance of agent $i_0$ with and without the feedback from the \system{}.  The evaluation metrics vary by domain as described below.

\begin{table*}[t]
    \centering
    {\small
\begin{tabular}{lccccccc}
    \toprule
    \multirow{2}{*}{Scenario} &  \multirow{2}{*}{\shortstack{Baseline \\(WGAN-GP)}}& \multicolumn{6}{c}{WGAN-GP-C over  $\beta\in[0,1]$}                \\\cline{3-8}\specialrule{0em}{0pt}{3pt}
                               &                           & 0 & 0.1 & 0.3 & 0.5 & 0.7 & 1.0 \\
    \midrule
    Bias-100                     &        0.705(0.009)	   &0.702(0.010)&0.655(0.016)&0.618(0.020)&0.585(0.010)&0.521(0.004)&0.483(0.014)     \\
    Bias-90                    &        0.650(0.011)	   &0.642(0.009)&0.630(0.016)&0.629(0.022)&0.610(0.030)&0.566(0.021)&0.473(0.002)     \\
    Low-32                &             0.641(0.006)				 &0.639(0.011)&0.614(0.008)&0.593(0.006)&0.586(0.004)&0.565(0.005)&0.541(0.004)     \\
    Low-64                &             0.582(0.010)	             &0.570(0.009)&0.567(0.012)&0.559(0.010)&0.552(0.007)&0.553(0.007)&0.539(0.008)     \\
\bottomrule
\end{tabular}
}
\caption{\MGM{} vs baseline in Wasserstein-1 distance (mean and std. dev. of 16 runs; each run is a sample from the final trained generator) for different scenarios and varying $\beta$.}
\label{table: s}
\end{table*}

\subsection{Synthetic Data}
We first show our results on synthetic data to explore various conditions under which the \MGM{} provides benefits. 
The synthetic data scenario models a system where the output of the \system{} is linearly dependent on the outputs of two agents. 
Agent 1 acts as the new arrival in this experiment. 

\textbf{Experimental Setup:} We generate 128,000 samples for all agents, output of agent 1, $x_1\in \mathbb{R}^2$, is drawn from Swiss roll (see Fig. \ref{fig: exp1-summary} (1)), and output of agent 2,  $x_2\in \mathbb{R}^2$, is drawn from a 2D Gaussian distribution. The data for the \system{} is a linear combination of agent outputs, $y = \beta x_1+(1-\beta) x_2$, where the parameter $\beta\in [0,1]$ controls the dependence of $y$ on $x_1$. We use Wasserstein-1 distance, computed by POT package \cite{flamary2021pot}, to evaluate the model. 


We use unconditional WGAN-GP to model the agents and conditional WGAN-GP to model the \system{}. Each WGAN-GP consists of a generator network with 3 fully-connected layers of 512 nodes with Leaky ReLU activations and a critic network of the same structure as the generator except for input and output layer.
We choose the weight of gradient penalty term $\lambda_{i_0}\!=\!0.1$ for agent GAN and $\lambda_{\sys{}}\!=\!1$ for \system{} GAN based on a grid search.
This experiment used the combined loss $\alpha L_a+(1-\alpha)L_f$, thus, 
the \MGM{} trained model is denoted as WGAN-GP-C in Table~\ref{table: s}.

\textbf{Biased data}: We remove the data of agent 1 from the top right area (see Fig.~\ref{fig: exp1-summary}(1)) either completely (bias-100) or 90\% (bias-90). We train with a minibatch size of 256. Then, the model is evaluated against a test set of size 2,000 that contains data in all area. 

\textbf{Low data}: We sample random subsets of the training data for agent 1, of size 32 and 64. We train with a minibatch size of 32 or 64, depending on the size of the training data. Then, same as the biased data scenario, the model is evaluated with a test set of size 2,000. 

\textbf{Results Discussion}: We visualize the trained distribution for bias-90 and low-32 ($\beta=0.7$) in Fig.~\ref{fig: exp1-summary} (2) and~\ref{fig: exp1-summary} (3) respectively. We report quantitative results in Table~\ref{table: s}. The results show that tighter coupling (higher $\beta$) results in better performance gain of agent 1's generator. This is consistent with the inference derived from our theoretical results (text after Lemma~\ref{lemma:dist}). \textcolor{black}{We observe that agent 1's generator improves more in the case where the learned distribution by a single GAN is further away from the true distribution. On average, low-32 improves 8.0\% compared with 4.4\% for low-64, while bias-100 improves 15.7\% compared with 9.0\% for bias-90, where a single WGAN-GP (baseline) performs worse for low-32 and bias-100. More results obtained on multiple different random low datasets can be found in appendix.}

\begin{table*}[t]
 \centering
{ \small
\begin{tabular}{lcccc}
    \toprule
     ($\times10^{-2}$)& Full  & Half  & Low-64  & Low-124  \\
    \midrule
        COT-GAN   &9.7(1.0)&13.9(1.6)&18.9(0.8)&10.2(1.1)\\
        COT-GAN-CJ&9.1(0.7)&12.6(1.0)&17.3(1.6)&\textbf{8.5(0.9)}\\
        COT-GAN-CM&\textbf{8.8(0.8)}&\textbf{12.2(0.9)}&\textbf{12.8(1.4)}&9.4(0.9)\\
    \bottomrule
    \end{tabular}
    }
\caption{\textcolor{black}{Model comparison using ACE ($\downarrow$) (avg and std. dev. of 16 sample means, each of 16 runs; each run is a sample from the final trained generator) for different scenarios. Bold shows best results.}}
\label{table: ts}
\end{table*}

\begin{table}[t]
\centering
{\small
\begin{tabular}{lccc}
    \toprule
          & 5-Shot & 10-Shot & 30-Shot \\
    \midrule
FSGAN     &60.01(0.47)&49.85(0.71)&\textbf{47.64(0.69)}\\
FSGAN-A   &\textbf{56.28(1.14)}&\textbf{48.98(0.52)}&48.41(1.10)\\
FreezeD   &78.07(0.95)&53.69(0.38)&47.94(0.87)\\
FreezeD-A &\textbf{77.61(1.78)}&\textbf{51.92(0.73)}&\textbf{46.27(0.56)}\\
    \bottomrule
\end{tabular}
}
\caption{FID scores ($\downarrow$) where \MGM{} uses alternate updating. Std. dev. are computed across 5 runs.}
\label{table: fs}
\end{table}

\begin{table*}[t]
    \centering
{\small
\begin{tabular}{lccccccc}
    \toprule
    \multirow{2}{*}{Scenario} &  \multirow{2}{*}{\shortstack{Baseline \\(WGAN-GP)}}& \multicolumn{6}{c}{WGAN-GP-A over  $\beta\in[0,1]$}                \\\cline{3-8}\specialrule{0em}{0pt}{3pt}
                               &                           & 0 & 0.1 & 0.3 & 0.5 & 0.7 & 1.0 \\
    \midrule
    Bias-100                     &        0.705(0.009)	   &0.735(0.008)&0.664(0.018)&0.622(0.018)&0.603(0.018)&0.591(0.018)&0.488(0.008)    \\
    Low-64                &             0.582(0.010)				  &0.571(0.007)&0.571(0.008)&0.575(0.005)&0.559(0.006)&0.551(0.004)&0.551(0.007)     \\
\bottomrule
\end{tabular}
}
\caption{Ablation study, \MGM{} vs baseline in Wasserstein-1 distance (over 16 runs) for 2 scenarios and varying $\beta$.}
\label{table: s-ablation}
\end{table*}

\subsection{Real-World Time Series Data}
Prior work has investigated synthesizing time series data of a single agent via GANs~\cite{cotgan2020xu, timegan2019yoon,esteban2017real}. We model the time series data of an agent who is interacting in an electricity market. 
The European electricity system enables electricity exchange among countries. The demand for the electricity in one country is naturally affected by the electricity prices in its neighboring countries within Europe because of the cross-border electricity trade. For example, the demand for electricity in Spain is affected by the electricity prices in France, Portugal, and Spain itself. 

We model the price in France as agent 1, the prices in Spain and Portugal as agents 2 and 3 respectively, and the demand in Spain as the \system{}. The intuition is that the dependency between the demand for electricity in Spain and the price in France is able to provide extra feedback for agent 1 and hence help with better modeling of agent 1. 

\textbf{Experimental Setup:} We use the electric price and demand data publicly shared by Red Eléctrica de España, which operates the national electricity grid in Spain~\cite{reees}. The data contains the day-ahead prices in Spain and neighboring countries, and the total electricity demand in Spain. We use the data between 5 Nov, 2018 and 13 July, 2019 aggregated over every hour and split by days. We use the difference between autocorrelation metric (ACE)~\cite{cotgan2020xu} to evaluate the output.

For each of 249 days, agent 1 outputs the price every hour in France, $x_1\in \mathbb{R}^{24}$, while other agents output the price every hour in Spain and Portugal, $x_2, x_3 \in \mathbb{R}^{24}$. The \system{} takes as input the prices in the three countries and outputs the total demand in Spain, $y \in \mathbb{R}^{24}$. We use COT-GAN \cite{cotgan2020xu} to model the agents and its conditional version to model the \system{}, using a minibatch size of 32, over 20,000 iterations. For other settings, we follow COT-GAN defaults.

As stated in preliminaries, when \emph{pre-training} the \system{}, the critic of the conditional COT-GAN should take as input the condition as well. However, we find that such pre-trained \system{} performs poorly. Instead, we try an \emph{alternate pre-training} where we do not feed the condition into the critic, thus, the critic measures the distance between the marginal distribution of the \system{} output and real data (discussed more in appendix); we denote the corresponding \MGM{} model by COT-GAN-CM (C for combined loss, M for marginal distribution). We denote the former \MGM{} model by COT-GAN-CJ (J for joint distribution). Empirical results show that COT-GAN-CM works better than COT-GAN-CJ.

\textbf{Biased data}: Modeling missing data in a biased manner is tricky in time series data. We take the simple approach of dropping the second half of the time series sequences, after which we have 124 time series sequences (call this Half) from Nov 2018 to Feb 2019, that may exhibit seasonality.

\textbf{Low data}: We randomly sample 64 and 124 time series sequences (out of the 249) once for agent 1 and treat these two subsets (low-64 and low-124) as examples of low data.

\textbf{Results Discussion}: The results are summarized in Table \ref{table: ts}. \textcolor{black}{We observe the two \MGM{} models consistently outperform the baseline and COT-GAN-CM outperforms COT-GAN-CJ most of the times.} The results provide evidence that our \MGM{} approach works for time series data. Additional supporting results are in the appendix.

\subsection{Few-Shot Image Generation}
The few-shot image generation problem is to generate images with very limited data, e.g., 10 images. Many works have proposed to solve this problem by adapting a pre-trained GAN-based model to a new target domain~\cite{mo2020freeze, ojha2021few, li2020few, wang2020minegan}. As low data is one of the scenarios in which our approach provides benefits, we explore few-shot image generation problem in a \MGM{} set-up.

We consider a single agent and a \system{}. 
The agent generates images of the \emph{miniature horses}, while the \system{} is a converter from horse to zebra images. The intuition here is that the \system{} can provide feedback about the high-level concept of a horse structure, such that it mitigates the overfitting of the agent due to limited training data. 

\begin{table}[t]
\centering
{\small
\begin{tabular}{lccc}
    \toprule
          & 5-Shot & 10-Shot & 30-Shot \\
    \midrule
FSGAN     &60.01(0.47)&49.85(0.71)&\textbf{47.64(0.69)}\\
FSGAN-C   &\textbf{55.79(0.82)}&\textbf{49.19(0.56)}&49.30(0.73)\\
FreezeD   &78.07(0.95)&\textbf{53.69(0.38)}&\textbf{47.94(0.87)}\\
FreezeD-C &\textbf{67.50(1.68)}&66.03(1.01)&69.45(0.45)\\
    \bottomrule
\end{tabular}
}
\caption{Ablation study, FID scores ($\downarrow$) where \MGM{} uses single combined loss. Std dev. are computed across 5 runs.}
\label{table: fs-ablation}
\end{table}

\textbf{Experiment Setup:} We collected images of miniature horses from the search engine \textit{yandex.com} by keyword ``miniature horse''. 
All the images were resized to $256 \times 256$. After processing, the dataset contains 1061 images, 34 of them forming training set, the remainder forming test set. The dataset will be released publicly. We employ the widely used FID~\cite{heusel2017gans} score to evaluate the output. 

The agent model produces images of miniature horse, $x$. The \system{} takes input $x$ to produce zebra images $y$, where $x,y\!\in\!\mathbb{R}^{256\times256\times3}$. 
We select two models, FSGAN~\cite{robb2020few} and FreezeD \cite{mo2020freeze} as the agent model; both of these fine tune a  pre-trained StyleGAN2~\cite{karras2020analyzing} to the miniature horse domain. We select the prior pre-trained CycleGAN~\cite{zhu2017unpaired} (horse to zebra) as the \system{} model. Recall that we train agent with $L_f$ followed by $L_a$ (alternate updating) in this experiment. We thus denote the \MGM{} version of FSGAN and FreezeD by FSGAN-A and FreezeD-A respectively.

We follow the defaults from FSGAN, using minibatch size of 16, stopping at iteration 1250 and choosing the better of the model at iteration 1000 or 1250. We evaluate the model in the same manner as \citet{ojha2021few}.

\begin{figure}[t]
    \centering
    \includegraphics[width=7.5cm,keepaspectratio]{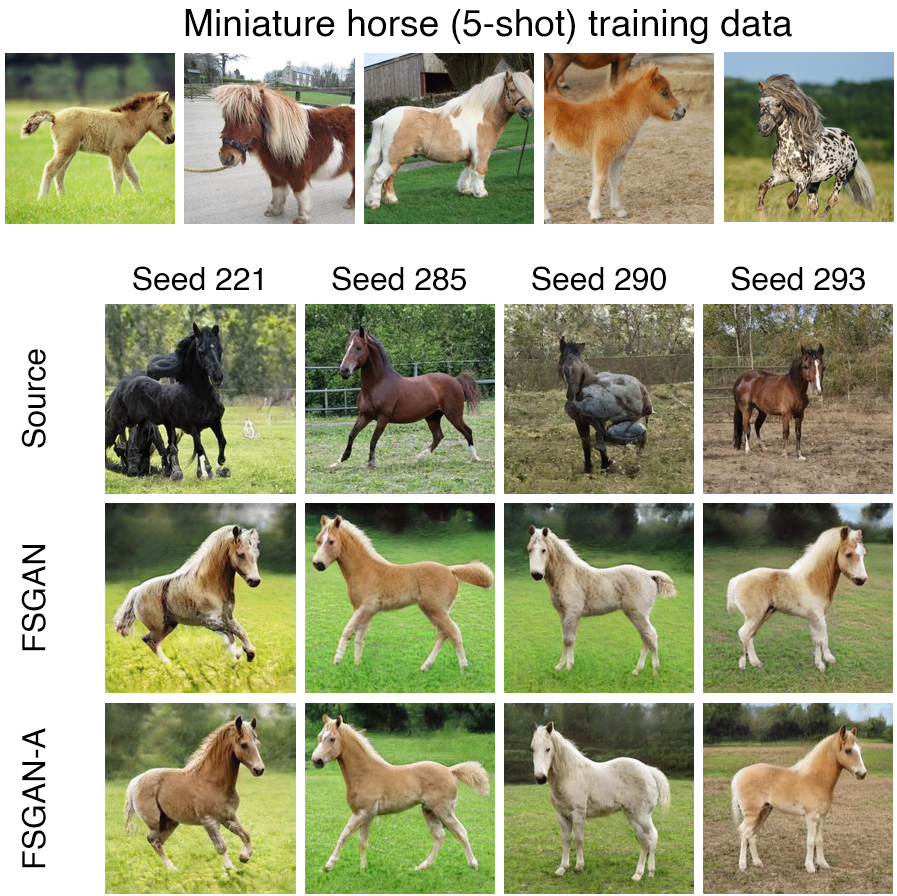}
    \caption{Source is the pretrained Style-GAN2, FSGAN is tuned Source, FSGAN-A is MGM tuned FSGAN.  Our \MGM{} approach (FSGAN-A) produces more natural images w.r.t. horse body, tail and clearer background.}
    \label{fig: exp3_sample}
\end{figure}

\textbf{Results Discussion}: The quantitative results in Table \ref{table: fs} with 5, 10, 30 shot for FSGAN and FreezeD show that the \MGM{} approach improves the baseline FSGAN and FreezeD, especially for lower shot settings. Also, we show some clear cases of improved quality of images in Fig.~\ref{fig: exp3_sample}.

\subsection{Ablation Study}
To further investigate how the \system{} improves the agent GAN model, 
we perform an ablation study for synthetic and image data. For synthetic data, instead of combined loss, we train the agent generator with $L_a$ and $L_f$ alternately. We denote this alternate by WGAN-GP-A and show results in Table~\ref{table: s-ablation}. We observe that the alternate updating strategy 
has less improvement than the combined loss (in Table~\ref{table: s}) in all cases, except low-64 with $\beta=0.7$. This reveals combined loss strategy is more effective for synthetic data. 
For image data, we perform ablation by using combined loss. We denote this alternate by FSGAN-C and FreezeD-C, and present results in Table~\ref{table: fs-ablation}. The results show that the alternate updating strategy is more effective on image data. Also, we found the training of FreezeD-C to be unstable, which is reflected in the results.

\section{Conclusion}


In this paper, we demonstrate that we can significantly improve the performance of generative models by incorporating real-world interactions. We presented and analyzed our \MGM{} approach both theoretically and empirically, and we show that our approach is particularly effective when data available is small. A possible future work could be to explore the approach with generative models other than GANs.

\section{Acknowledgement}
This research/project is supported by the National Research Foundation, Singapore under its AI Singapore Programme (AISG Award No: AISG2-RP-2020-017). We are thankful for the help from Tianlin Xu on COT-GAN. 

\bibliography{references}



\clearpage
\begin{bibunit}[aaai22]

\appendix
\renewcommand\thefigure{A.\arabic{figure}}
\renewcommand\thetable{A.\arabic{table}}
\setcounter{figure}{0}
\setcounter{table}{0}


\section{More Concrete Examples of the MGM Framework}
As stated in the main paper, the framework in Algorithm~\ref{alg:framework} can be instantiated with other GANs other than the WGAN presented in the main paper. Here we present instantiations with COT-GAN (as agent and \system{}), and StyleGAN2 as agent and CycleGAN as \system{}.

\addcontentsline{toc}{section}{Concrete Example of MGM Framework}
\textbf{Real-world time series generation}: The authors of the COT-GAN work proposed to generate time-series data by solving a casual optimal transport problem~\citep{cotgan2020xu}. The pipeline is (1) obtain the cost from the source distribution to the target distribution with the causality constraint, (2) compute a regularized distance between distributions using samples and use of the Sinkhorn algorithm (effect of the regularization is controlled by weight $\varepsilon$), (3) form the loss term by the regularized distances of last step. 

We first introduce some notation. There are two discriminator networks, denoted by $h_{w_1}: X \rightarrow \mathbb{R}^K$ and $M_{w_2}: X \rightarrow \mathbb{R}^K$ where $K$ is a constant, and let $w = (w_1, w_2)$. Then, the authors define and compute the cost $c_{w}^{\mathcal{K}}(x, \hat{x})$ by the equation
$$
c(x, \hat{x})+\sum_{k=1}^{K} \sum_{t=1}^{T-1} h_{w_{1}, t}^{k}(\hat{x}) \Delta_{t+1} M_{w_{2}}^{k}(x)
$$
where $c(\cdot,\cdot)$ is the Euclidean distance between the examples of $x$ and that of $\hat{x}$, $T$ is the length of the time series sequence, $\Delta_{t+1} M_{w_{2}}^{k}(\cdot)$ represents \textcolor{black}{$M^k_{w_2, t+1}(\cdot) - M^k_{w_2, t}(\cdot)$}. The cost $c_{w}^{\mathcal{K}}(x, \hat{x})$ represents the underlying distance between two time series in the space of time series. For more details of the form of the cost, refer to the paper~\citep{cotgan2020xu}.

The authors also propose a regularized COT distance between causal couplings over time-series (Eq.~3 in~\citep{cotgan2020xu}), which is an expectation. The empirical version of the regularized distance ${W}_{c_w,\varepsilon}(\mathbf{x}, \hat{\mathbf{x}})$ can be approximated by the Sinkhorn algorithm, where $\mathbf{x}$ is an empirical distribution formed from samples of a mini-batch. 

However, the authors show that the simple application of Sinkhorn algorithm with standard Sinkhorn divergence provides a poor estimate of the regularized distance. Hence, the authors defined a mixed Sinkhorn divergence, which uses two sets of samples for each of real data and generated data as a variance reduction technique. Mixed Sinkhorn divergence $\widehat{{W}}_{c, \varepsilon}^{\operatorname{mix}}\left(\mathbf{x}, \mathbf{x}^{\prime}, {\hat{\mathbf{x}}}_{\theta}, \hat{{\mathbf{x}}}_{\theta}^{\prime}\right)$ is given as:
$$
{W}_{c, \varepsilon}\left({\mathbf{x}}, \hat{\mathbf{x}}_{\theta}\right)+
{W}_{c, \varepsilon}\left({\mathbf{x}}^{\prime}, \hat{\mathbf{x}}_{\theta}^{\prime}\right)-
{W}_{c, \varepsilon}\left({\mathbf{x}}, {\mathbf{x}}^{\prime}\right)-
{W}_{c, \varepsilon}\left(\hat{\mathbf{x}}_{\theta}, \hat{\mathbf{x}}_{\theta}^{\prime}\right)
$$
where $\hat{\mathbf{x}}_\theta$ is the empirical distribution from one mini-batch generated by the generator, $\hat{\mathbf{x}}_\theta^{\prime}$ is the empirical distribution from another mini-batch generated by the generator, $\mathbf{x}^{\prime}$ is the empirical distribution from one mini-batch of real data and $\mathbf{x}'$ is the empirical distribution from another mini-batch of real data. (Note that this algorithm requires two mini-batches to compute $\widehat{{W}}_{c, \varepsilon}^{\operatorname{mix}}$). When updating the discriminator network, the author introduced a regularization term $p_{\mathbf{M}_{w_{2}}}(x)$. 

We utilize the COT-GAN in our \MGM{} set-up as follows: $L_a$ and $L_f$ is $\widehat{{W}}_{c_{w_{i_0}}, \varepsilon}^{\operatorname{mix}}\left(\mathbf{x}, \mathbf{x}^{\prime}, {\hat{\mathbf{x}}}_{\theta_{i_0}}, \hat{{\mathbf{x}}}_{\theta_{i_0}}^{\prime}\right)$ and $\widehat{{W}}_{c_{w_{mix}}, \varepsilon}^{\operatorname{mix}}\left(\mathbf{y}, {{\mathbf{y}^{\prime}}}, {\hat{\mathbf{y}}}_{\theta_{mix}}, \hat{{\mathbf{y}}}_{\theta_{mix}}^{\prime}\right)$ respectively. Then, $W_a$ is
$$
- L_a + \lambda p_{\mathbf{M}_{w_{2,i_0}}}(\mathbf{x}_{i_0})
$$
and $W_f$ is 
$$
- L_f + \lambda p_{\mathbf{M}_{w_{2,mix}}}(\mathbf{y})
$$

\textbf{Few-shot image generation}: We use StyleGAN2 as the agent and CycleGAN as the \system{}. 
The $\mathcal{L}_f$ is $\frac{1}{2}{E}_{y\sim\mathbb{P}^{g}_{y}}[(C_{w_{mix}}(y)-c)^2]$. $\mathcal{W}_f$ is the LSGNA loss~\citep{mao2017least}
$$
\frac{1}{2} {E}_{y\sim\mathbb{P}^r_y}[(C_{w_{mix}}(y)-b)^2]
+ \frac{1}{2} {E}_{y\sim\mathbb{P}^g_y}[(C_{w_{mix}}(y)-a)^2]
$$
where we use $b=c=1$, $a=0$

For expected loss $\mathcal{L}_a$, the author of StyleGAN2 introduced a new regularizer, path length regularization, denoted by $\mathcal{J}^{PL}_{G_{\theta}}$. The whole expected loss $\mathcal{L}_a$ is
$$
- {E}_{x_{i_0}\sim\mathbb{P}^g_{x_{i_0}}}[\log C_{w_{i_0}}(x_{i_0})]
+ \lambda\mathcal{J}^{PL}_{G_{\theta_{i_0}}}
$$
$\mathcal{W}_a$ is the standard GAN loss with $R_1$ regularization, denoted by $\mathcal{R}$

\begin{align*}
    &- {E}_{x_{i_0}\sim\mathbb{P}^r_{x_{i_0}}}[\log C_{w_{i_0}}(x_{i_0})]
    - {E}_{x_{i_0}\sim\mathbb{P}^g_{x_{i_0}}}[\log(1- C_{w_{i_0}}(x_{i_0}))]\\
    &+ \lambda \mathcal{R}_{C_{w_{i_0}}, \mathbb{P}^r_{x_{i_0}}}
\end{align*}

Note these two models are pre-trained models. We use the online implementation
\footnote{StyleGAN2: https://github.com/NVlabs/stylegan2\\ CycleGAN: https://github.com/XHUJOY/CycleGAN-tensorflow}. 
The loss terms thus follow that in their implementation.
In different GANs, people denote the discriminator net by $D_w$, $C_w$ or $F_w$ depending on the tasks. For ease of presentation, we use $C_w$ in all cases.


\section{More Experiment Results of Synthetic Data}
Due to the randomness of sampling low datasets, the results vary for different low datasets. However, \MGM{} consistently shows effective improvement compared with the baseline. We show results on different low datasets of size 32 and size 64 in Table \ref{table: s-appendix}.

\begin{table*}[t]
    \centering
{\small
\begin{tabular}{lcccccccc}
    \toprule
    \multirow{2}{*}{Scenario} &\multirow{2}{*}{Seed}&  \multirow{2}{*}{\shortstack{Baseline \\(WGAN-GP)}}& \multicolumn{6}{c}{WGAN-GP-C over  $\beta\in[0,1]$}                \\\cline{4-9}\specialrule{0em}{0pt}{3pt}
                               &    &                       & 0 & 0.1 & 0.3 & 0.5 & 0.7 & 1.0 \\
    \midrule
    Low-32      &6128& 0.660(0.010)&0.670(0.011)&0.619(0.008)&0.611(0.010)&0.601(0.013)&0.588(0.018)&0.531(0.008)    \\
    Low-32      &968 & 0.607(0.009)&0.607(0.010)&0.597(0.006)&0.592(0.007)&0.584(0.007)&0.538(0.004)&0.539(0.009)    \\
    Low-32      &1025& 0.572(0.007)&0.564(0.008)&0.573(0.011)&0.555(0.011)&0.536(0.003)&0.536(0.004)&0.510(0.004)    \\
    Low-64      &6160& 0.558(0.009)&0.542(0.009)&0.534(0.008)&0.525(0.006)&0.518(0.007)&0.538(0.006)&0.528(0.011)    \\
    Low-64      &3786& 0.552(0.007)&0.551(0.009)&0.542(0.007)&0.536(0.005)&0.519(0.006)&0.538(0.009)&0.500(0.010)    \\
    Low-64      &3304& 0.545(0.006)&0.535(0.006)&0.538(0.008)&0.536(0.010)&0.531(0.006)&0.531(0.004)&0.518(0.007)    \\
\bottomrule
\end{tabular}
}
\caption{MGM vs baseline in Wasserstein-1 distance (mean and std. dev. of 16 runs) for different low datasets and varying $\beta$.}
\label{table: s-appendix}
\end{table*}

\section{Training of the Mixer for COT-GAN}
\addcontentsline{toc}{section}{Training of the Mixer for COT-GAN}
One of the requirements to make the \MGM{} approach work is a good pre-trained \system{} generator. Generally, we can obtain such a \system{} generator by the manner introduced in preliminaries. However, for real world time series data, we observed poor performance of the conditional COT-GAN when the critic of the conditional COT-GAN takes as input the conditions. 


First, we argue why the mixed Sinkhorn divergence $\widehat{\mathcal{W}}_{c, \varepsilon}^{\operatorname{mix}}\left({\mathbf{x}}, \mathbf{x}^{\prime}, {\hat{\mathbf{x}}}_{\theta}, \hat{{\mathbf{x}}}_{\theta}^{\prime}\right)$ is not the correct term to use for a conditional version. Here every sample in a mini-batch is a condition time-series followed by the continuation time-series (for fake one also the condition is from real data and then the generator produces fake continuation). Note that when samples are generated for the different mini-batches $\mathbf{x}, {\mathbf{x}}^{\prime}, \hat{\mathbf{x}}_{\theta} , \hat{\mathbf{x}}_{\theta}^{\prime}$ then the conditions within real mini-batches $\mathbf{x}$ and $ {\mathbf{x}}^{\prime}$ are different and also within fake mini-batches $\hat{\mathbf{x}}_{\theta}$ and $ \hat{\mathbf{x}}_{\theta}^{\prime}$. Comparing these terms with different conditions is meaningless as the conditional distribution is dependent on the condition. Thus, mixed Sinkhorn divergence does not apply here.

\begin{figure*}[t]
    \centering
    \includegraphics[width=15cm,keepaspectratio]{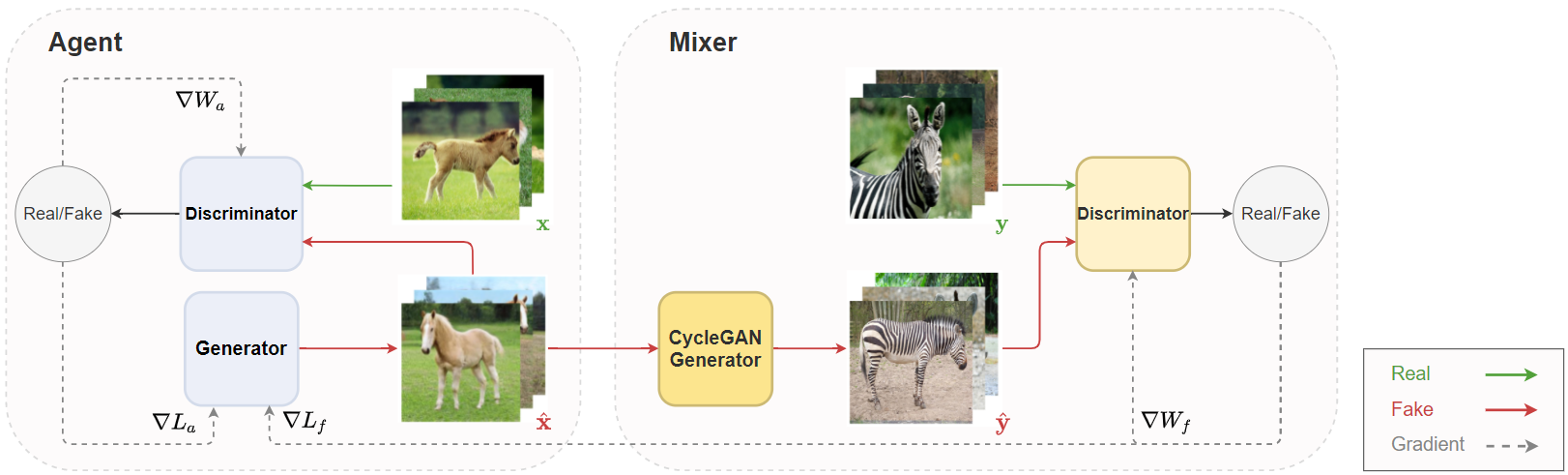}
    \caption{A specific instance of \MGM{} on few-shot image generation}
    \label{fig: arch-exp3}
\end{figure*}

As a consequence we train by not feeding conditions into the critic, which still learns quite well. We explain this in easier set-up using WGAN and the notation in preliminaries. Feeding conditions into the critic measures the Wassertein-1 distance between a joint distributions, $W(\mathbb{P}^r_{joint}, \mathbb{P}^g_{joint})$, which provides strong supervision to the generator, while not feeding conditions into the critic measures the distance between marginal distributions, $W(\mathbb{P}^r_{y}, \mathbb{P}^g_{y})$, marginalized over condition $c$. When we learn a conditional generator by minimizing $W(\mathbb{P}^r_{y}, \mathbb{P}^g_{y})$ with batch optimization, the generator can ignore the relationship between $c$ and $y$. However, the nature of mini-batch optimization prevents the generator from ignoring this relation. In a mini-batch, the empirical marginal distribution of $y$ is different among mini-batches (due to limited size of mini-batches) and this highly depends on the corresponding conditions (which are not fed to critic but fed to generator). The generator can do best by taking into account $c$ to minimize the distance between the real mini-batch distribution and the generated mini-batch distribution, especially for continuous sample-condition pair $(c, y)$ as is the case of our real time series experiment.

Overall, we pre-train the \system{} generator of COT-GAN by not feeding conditions into the critic. We empirically observed the better performance of this approach against the approach that uses condition with mixed Sinkhorn divergence (as we argued, mixed Sinkhorn divergence is meaningless to start with in conditional case). This empirical observation is shown on Table~\ref{table: cond_cot}. 

As the work~\citep{cotgan2020xu} shows that standard Sinkhorn divergence also provides poor estimates, thus, the design of a conditional COT-GAN where the critic explicitly takes as input the condition is an open question. In any case, these results actually provide more support for our \MGM{} approach.
The result in Table~\ref{table: cond_cot} and Table~\ref{table: ts} together supports our claim that a good \system{} generator is critical for the \MGM{} approach.

\begin{table}[t]
\centering
\begin{tabular}{lcc}
           \toprule
           & Mean(std) & Max \\
           \midrule
CMIX-COT-GAN-J &0.958(0.458)&2.583\\
CMIX-COT-GAN-M &\textbf{0.413(0.266)}&\textbf{1.341}\\
           \bottomrule
\end{tabular}
\caption{We compare the conditional COT-GAN with and without feeding conditions into the critic, denoted by CMIX-COT-GAN-J and CMIX-COT-GAN-M respectively. The results are in the ACE ($\downarrow$) metric (avg (std. dev.) and maximum over 100 runs with a batch size of 64) between the real time series and the generated time series given the \emph{same condition} in each batch.  Bold shows best results.}
\label{table: cond_cot}
\end{table}

\section{Continuous Condition for Conditional WGAN-GP}
\addcontentsline{toc}{section}{Continuous Condition for Conditional WGAN-GP}
The gradient regularizer $\mathcal{J}_{non}$ of non-conditional WGAN-GP~\citep{gulrajani2017improved} is
$$
\underset{y \sim \mathbb{P}_{y}^{\tilde r}}{\mathbb{E}}\left[\left(\left\|\nabla_{{{y}}} C({{y}})\right\|_{2}-1\right)^{2}\right]
$$
where $\mathbb{P}_{y}^{\tilde r}$ is is a mixture of data distribution $\mathbb{P}_{y}^{r}$ and the generated distribution $\mathbb{P}_{y}^{g}$. The $\mathcal{J}_{non}$ is a way to enforce the Lipschitz constraint by penalizing the gradient w.r.t $y\sim \mathbb{P}^{\tilde{r}}_y$. 
As we state in preliminaries, the critic of the conditional WGAN-GP estimates the distance between the joint distributions, $W(\mathbb{P}^r_{joint}, \mathbb{P}^g_{joint})$. Theoretically, the regularizer should penalize the gradient w.r.t $(y,c) \sim \mathbb{P}^{\tilde{r}}_{joint}$. However, we did not find any implementation of such a version online. 

A variation of $\mathcal{J}_{disc}$ in WGAN-GP~\citep{zheng2020conditional} conditioned on the discrete conditions, e.g. labels, is
$$
\underset{(y, c) \sim \mathbb{P}_{joint}^{\tilde r}}{\mathbb{E}}\left[\left(\left\|\nabla_{{{y}}} C({{y,c}})\right\|_{2}-1\right)^{2}\right]
$$
They ignore the condition $c$ when they compute $\mathcal{J}_{disc}$ as discrete labels have no gradients. 

In the experiment on synthetic data, the conditions are continuous. So we use this penalty term $\mathcal{J}_{cont}$
$$
\underset{(y, c) \sim \mathbb{P}_{joint}^{\tilde r}}{\mathbb{E}}\left[\left(\left\|\nabla_{{{y,c}}} C({{y,c}})\right\|_{2}-1\right)^{2}\right]
$$
And we find $\mathcal{J}_{cont}$ works better than $\mathcal{J}_{disc}$ in the case of continuous conditions.

\section{Architecture Diagram}
\addcontentsline{toc}{section}{Architecture Diagram}
We provide the architecture (see Fig.~\ref{fig: arch-exp3}) of the \MGM{} set-up that we used for the image domain, to illustrate the transfer learning nature of the \MGM{} set-up and the capability of the \MGM{} model on the few-shot learning task.

\section{Missing Proofs}
\addcontentsline{toc}{section}{Missing Proofs}
\subsection{Proof of Theorem~\ref{thm:main}}
\begin{proof}
Let $\pi^*_d(x_j, x_j')$ denote the distribution (or coupling) that achieves the min for the Wasserstein-1 distance below between the same distributions $\mathbb{P}^r_j$ and $\mathbb{P}^r_j$ for $j \neq i_0$
$$
\pi^*_d =  \arginf\limits_{\pi \in \Pi_{\mathbb{P}^r_j,\mathbb{P}^r_j}} \int^{}_{X_j \times X_j} d(x_j,x'_j) \; d\pi(x_j,x_j') 
$$
By definition, $ W(\mathbb{P}^r_j, \mathbb{P}^r_j) = 0$ for $j \neq i_0$.

Also, there exists functions $f^*_{j,d}$ and $g^*_{j,d}$ with $f^*_{j,d}(x_j) + g^*_{j,d}(x_j') \leq d(x_j, x_j')$ such that
\begin{align} 
\nonumber 0& = W(\mathbb{P}^r_j, \mathbb{P}^r_j)  = \int^{}_{X_j \times X_j}  d(x_j,x'_j) \; d\pi^*_d(x_j,x_j') \\
\nonumber &= \int^{}_{X_j} f^*_{j,d}(x_j) P^r_j(x_j) dx_j +  \int^{}_{X_j} g^*_{j,d}(x_j) P^r_j(x_j) dx_j \\
\nonumber & = \sup_{\substack{f_j(x_j) + g_j(x_j') \leq d(x_j, x_j') \\ f_j, g_j \in \mathcal{C}_b(X_j)}} \\
\nonumber & \qquad \int^{}_{X_j} f_j(x_j) P^r_j(x_j) dx_j +  \int^{}_{X_j} g_j(x_j) P^r_j(x_j) dx_j
\end{align}
Note that $f^*,g^*,\pi^*$ take $d$ as a subscript, denoting that they depend on the distance metric $d$. We denote this explicitly since we later deal with different distance metrics. Also, the notation $\pi^*$ is abused and used for every $j$ but the usage will be clear from the arguments of $\pi^*$.
Also, $0 = \int^{}_{X_j \times X_j}  d(x_j,x'_j) d\pi^*_d(x_j,x_j')$ and the fact that $d$ is distance ($d(x_j,x_j') =0$ iff $x_j = x_j'$) implies that for $j \neq i_0$
\begin{align}
    \nonumber\int^{}_{X_j \times X_j}  h(x_j,x'_j) \; d\pi^*_d(x_j,x_j') = \int^{}_{X_j }  h(x_j,x_j)  P^r_j(x_j) \; dx_j\\
    = E_{x_j \sim \mathbb{P}^r_j} [h(x_j, x_j)]\label{eq:split}
\end{align} 
for any measurable function $h$. This can be proved by showing that for the event $E$ defined as $\{x_j \neq x'_j\}$ we must have $\pi^*_d(x_j,x_j') (E) = 0$ in order to obtain $0 = \int^{}_{X_j \times X_j}  d(x_j,x'_j) d\pi^*_d(x_j,x_j')$.

Very similarly, let $\pi^*_{d_0}(x_{i_0}, x_{i_0}')$ denote the joint distribution that achieves the min for the Wasserstein-1 distance below between the different densities $P^r_{i_0}$ and $P^g_{i_0}$.
$$
\pi^*_{d_0} = \arginf\limits_{\pi \in \Pi_{P^r_{i_0},P^g_{i_0}}} \int^{}_{X_{i_0} \times X_{i_0}} d(x_{i_0},x'_{i_0}) \; d\pi(x_{i_0},x_{i_0}')  
$$
Again by definition of couplings, $\int_{X \times A}  d\pi^*_{d_0}(x_{i_0}, x_{i_0}') = \int_{A} P^g_{i_0}(x_{i_0}') dx_{i_0}'$ and $\int_{A \times X} d\pi^*_{d_0}(x_{i_0}, x_{i_0}')  = \int_A P^r_{i_0}(x_{i_0}) dx_{i_0}$ (here we used the assumption that density exists)

Further, let $\pi^*_{\hat{x},\hat{x}',d}(y, y')$ denote the joint distribution that achieves the min for the Wasserstein-1 distance below between $P^{r}_{y|\hat{x}}$ and $P^{r}_{y|\hat{x}'}$ for any fixed $\hat{x},\hat{x}'$.
$$
W(\mathbb{P}^{r}_{y|\hat{x}}, \mathbb{P}^{r}_{y|\hat{x}'})  = \inf_{\pi \in \Pi_{\mathbb{P}^{r}_{y|\hat{x}},\mathbb{P}^{r}_{y|\hat{x}'}} }
\int^{}_{Y \times Y} d(y,y') \; d\pi (y,y') 
$$
Similar functions $f^*_{\hat{x},d}$ and $g^*_{\hat{x}',d}$ (as for the earlier cases of $x$) exists here such that $f^*_{\hat{x},d}(y) + g^*_{\hat{x}',d}(y') \leq d(y,y')$ for all $y,y'$ as well as for any $\hat{x},\hat{x}'$.

Then, define the probability distribution
\begin{align*}
& \pi^*(y, y')(E) = \\
& \int_{X\times X} \mathbf{1}_{E} \; d\pi^*_{x,x',d}(y, y') d\pi^*_{d_0}(x_{i_0}, x_{i_0}') \prod_{j\neq i_0} d\pi^*_d(x_j, x_j') 
\end{align*}
for any event $E$ in the appropriate $\sigma$-algebra ($\mathbf{1}_{E}$ is the indicator for $E$). We will define $d_0$ later, which for now is any arbitrary distance metric. We want to show that the distribution $\pi^*(y, y')$ is the minimizer for 
$$
W(\mathbb{P}^r_y, \mathbb{P}^g_y) = \inf_{\pi \in \Pi_{P^r_y, P^g_y}}\int_{Y\times Y'} d(y,y') \; d\pi(y, y')
$$

Consider 
$$
\int_{Y\times Y'} d(y,y') \; d\pi^*(y, y')
$$
By definition, this is same as
$$
\int_{Y\times Y'} d(y,y')  \; \int_{X\times X'} \; d\pi^*_{x,x',d}(y, y') d\pi^*_{d_0}(x_{i_0}, x_{i_0}') \prod_{j \neq i_0} d\pi^*(x_j, x_j')
$$
This is same as (using Fubini's theorem~\citep{billingsley2008probability})
\begin{align*}
&\int_{Y\times Y' \times X\times X'} \Big [d(y,y')  \; d\pi^*_{x,x',d}(y, y') \\
& \qquad\qquad\qquad\qquad d\pi^*_{d_0}(x_{i_0}, x_{i_0}') \prod_{j \neq i_0} d\pi^*_d(x_j, x_j') \Big]
\end{align*}
Grouping the first two terms that have $y,y'$ and integrating over $Y \times Y'$ (using Fubini's theorem) we get the above term is same as
$$
\int_{X\times X} W(\mathbb{P}^{r}_{y|x}, \mathbb{P}^{r}_{y|x'})  \; d\pi^*_{d_0}(x_{i_0}, x_{i_0}') \prod_{j \neq i_0} d\pi^*_d(x_j, x_j')
$$
Using dual representation of the Wasserstein distance this is same as
\begin{align*}
\int_{X\times X'} & \Big[ \big ( \int_Y f^*_{x,d}(y) P^{r}_{y|x}(y|x) \; dy  + \int_Y g^*_{x',d}(y) P^{r}_{y|x'}(y|x') \; dy \big) \\
& \quad d\pi^*_{d_0}(x_{i_0}, x_{i_0}') \prod_{j \neq i_0} d\pi^*(x_j, x_j') \Big ]
\end{align*}

As $\pi^*$'s are couplings, we have the fact that $\int_{X \times X'} d\pi^*_{d_0}(x_{i_0}, x_{i_0}') \prod_{j \neq i_0} d\pi^*(x_j, x_j')  = \int_{X} P^r_{i_0}(x_{i_0})\prod_{j \neq {i_0}} P^r_j(x_j) dx$ (from marginal properties of coupling $\pi^*$ and existence of densities) and very similarly $\int_{X \times X'} \pi^*_{d_0}(x_{i_0}, x_{i_0}') \prod_{j \neq i_0} \pi^*(x_j, x_j') dx = \int_{X'} P^g_{i_0}(x'_{i_0})\prod_{j \neq {i_0}} P^r_j(x'_j) dx'$.

Since $X'$ is same as $X$, and plugging back in 
\begin{align}
\nonumber &= \int_{X} \int_Y f^*_{x}(y) P^{sys}_{y|x}(y|x) P^r_{i_0}(x_{i_0})\prod_{j \neq {i_0}} P_j(x_j) \; dy \; dx \; + \\
\nonumber & \qquad \int_{X} \int_Y g^*_{x'}(y) P^{sys}_{y|x'}(y|x') P^g_{i_0}(x'_{i_0})\prod_{j \neq {i_0}} P_j(x'_j)  \; dy \; dx'  \\
\nonumber & = \int_{X} \int_Y f^*_{x}(y) P^r_{joint}(y,x) \; dy \; dx \; + \\
& \qquad\qquad \int_{X} \int_Y g^*_{x}(y) P^g_{joint}(y,x)  \; dy \; dx \label{eq:joint}
\end{align}

Define $F(y) = \int_{X}  f^*_{x}(y) P^r_{joint}(y,x)  \; dx \Big / P^r_y(y)$ and $G(y) = \int_{X}  g^*_{x}(y) P^g_{joint}(y,x)  \; dx \Big / P^g_y(y)$ for $y \in A$ and $y \in B $ respectively. First, by assumption $P^r_y(y) > 0 $ for all $y \in Y$ and $P^r_y(\cdot)$ is continuous. 

We claim that $\int_{X}  f^*_{x}(y) P^r_{joint}(y,x)  \; dx$ is also continuous in $y$. First, $f^*_{x}(y) P^r_{joint}(y,x)$ is jointly continuous in $x,y$ since we assumed that $f^*_{x}(y)$ is jointly continuous in $x,y$ and all densities are continuous. As the space $X \times Y$ is compact, $f^*_{x}(y) P^r_{joint}(y,x)$ is also bounded by Weierstrass extreme value theorem~\citep{rudin1986principles}.
Then, define $h_n(x) = f^*_{x}(y_n) P^r_{joint}(y_n,x)$ for some sequence $y_n$ converging to $y$ and clearly $h_n(x)$ is continuous and bounded for every $n$. Then, we can apply  dominated convergence theorem~\citep{billingsley2008probability} to get
$$
\lim_{y_n \rightarrow y} \int_{X}  h_n(x)  \; dx = \int_{X}  \lim_{y_n \rightarrow y} h_n(x)  \; dx
$$
The above yields
$$
\lim_{y_n \rightarrow y} \int_{X}  h_n(x)  \; dx = \int_{X}  f^*_{x}(y) P^r_{joint}(y,x) \; dx
$$
which proves our claim.

Next, since $F, G$ are continuous in $y$ and $Y$ is compact, we have $F, G$ are bounded by Weierstrass extreme value theorem. Hence, $F, G \in \mathcal{C}_b(Y)$
Thus, Equation~\ref{eq:joint} can be written as
$$
\int_Y F(y) P^r_y(y) \; dy  + \int_Y G( y) P^g_y(y) \; dy 
$$

Overall, we proved that
\begin{align*}
W( & \mathbb{P}^r_y, \mathbb{P}^g_y)  = \inf_{\pi \in \Pi_{P^r_y, P^g_y}}\int_{Y\times Y'} d(y,y') \; d\pi(y, y')\\
& \leq \int_{Y\times Y'} d(y,y') \; d\pi^*(y, y')\\
& = \int_{X\times X} W(\mathbb{P}^{r}_{y|x}, \mathbb{P}^{r}_{y|x'})  \; d\pi^*_{d_0}(x_{i_0}, x_{i_0}') \prod_{j \neq i_0} d\pi^*_d(x_j, x_j')\\
& = \int_Y F(y) P^r_y(y) \; dy  + \int_Y G( y) P^g_y(y) \; dy \\
& \leq \sup_{\substack{f(y) + g(y') \leq d(y,y') \\ f,g \in \mathcal{C}_b(Y)}}\int_Y f(y) P^r_y(y) \; dy  + \int_Y g( y) P^g_y(y) \; dy \\
& = W(\mathbb{P}^r_y, \mathbb{P}^g_y)
\end{align*}

Using Equation~\ref{eq:split} we can infer that $$\int_{X\times X} W(\mathbb{P}^{r}_{y|x}, \mathbb{P}^{r}_{y|x'})  \; d\pi^*_{d_0}(x_{i_0}, x_{i_0}') \prod_{j \neq i_0} d\pi^*_d(x_j, x_j')$$ is same as
\begin{align*}
&\int_{X_{i_0} \times X_{i_0}} E_{x_{-i_0} \sim \otimes_{j \neq i_0} \mathbb{P}^r_j}\Big[ \\
& \qquad\quad W(\mathbb{P}^{r}_{y| \langle x_{-i_0}, x_{i_0} \rangle}, \mathbb{P}^{r}_{y|\langle x_{-i_0}, x'_{i_0} \rangle }) \Big]   \; d\pi^*_{d_0}(x_{i_0}, x_{i_0}') 
\end{align*}
The above was proven without any assumption on the metric $d_0$ (or even on the metric for $Y$ or $X_j$'s), hence it holds for any metric.

\end{proof}

\subsection{Proof of Lemma~\ref{lemma:dist}}
\begin{proof}
We first claim that if $M_{x_{-i_0}}(\cdot )$ is an injection (one-to-one) for some $x_{-i_0}$, then 
$${\small 
\tilde{d}(x_{i_0}, x'_{i_0}) = W(\mathbb{P}^{r}_{y| \langle x_{-i_0}, x_{i_0} \rangle }, \mathbb{P}^{r}_{y|\langle x_{-i_0}, x'_{i_0} \rangle })
}$$
is a distance metric on $X_{i_0}$ for that $x_{-i_0}$.
It is a known result that if $M: X \rightarrow Y$ is an injection and $d_Y$ is a distance metric on $Y$ then $d$ defined as $d(x_1, x_2) = d_Y (M(x_1), M(x_2))$ is a metric on $X$. Also, if $M$ is not an injection then $d_Y$ is a pseudo-metric. This directly proves the claim above about $\tilde{d}$, where the Wasserstein-1 metric is the analog to $d_Y$ here. 

For, the second claim, some parts follow easily: (1) $d_0(x_{i_0}, x'_{i_0}) = d_0(x'_{i_0}, x_{i_0})$ is by definition, (2) triangle inequality follows from monotonicity of expectations and the triangle inequality of the term inside the expectation, and (3) $d_0(x_{i_0}, x_{i_0}) = 0$ is by definition. This already shows the pseudo-metric nature of $d_0$. The only tricky part is showing that $d_0(x_{i_0}, x'_{i_0}) = 0$ implies $x_{i_0} = x'_{i_0}$ in case of the injection. For this, let $d_0(x_{i_0}, x'_{i_0}) = 0$ which is same as
$$
 E_{x_{-i_0} \sim \otimes_{j \neq i_0} \mathbb{P}^r_j}\big[ W(\mathbb{P}^{r}_{y| \langle x_{-i_0}, x_{i_0} \rangle}, \mathbb{P}^{r}_{y| \langle x_{-i_0}, x'_{i_0} \rangle}) \big ] = 0
$$
This implies that $W(\mathbb{P}^{r}_{y| \langle x_{-i_0}, x_{i_0} \rangle}, \mathbb{P}^{r}_{y| \langle x_{-i_0}, x'_{i_0} \rangle }) = 0$ almost everywhere w.r.t. the measure $\otimes_{j \neq i_0} \mathbb{P}^r_j$. For contradiction, assume $x_{i_0} \neq x'_{i_0}$. By assumption of injection, $\mathbb{P}^{r}_{y| \langle x_{-i_0}, x_{i_0} \rangle}$ and $ \mathbb{P}^{r}_{y| \langle x_{-i_0},x_{i_0} \rangle}$ are different distributions. Hence, $W(\mathbb{P}^{r}_{y| \langle x_{-i_0}, x_{i_0} \rangle}, \mathbb{P}^{r}_{y| \langle x_{-i_0}, x'_{i_0} \rangle }) > 0$ for all $x_{-i_0}$ but this contradicts the inferred result that $W(\mathbb{P}^{r}_{y| \langle x_{-i_0}, x_{i_0} \rangle }, \mathbb{P}^{r}_{y| \langle x_{-i_0}, x'_{i_0} \rangle }) = 0$ almost everywhere w.r.t. the measure $\otimes_{j \neq i_0} \mathbb{P}^r_j$. Hence, we must have $x_{i_0} = x'_{i_0}$.
This proves all properties required to make $d_0$ a distance metric on $X_{i_0}$ in case of the injection.
\end{proof}
\putbib[references]
\end{bibunit}

\end{document}